\documentclass[10pt,twocolumn,letterpaper]{article}

\usepackage[pagenumbers]{iccv} 

\usepackage{amsfonts}       
\usepackage{graphicx}
\usepackage{subcaption}
\usepackage{amsmath}
\usepackage{amssymb}
\usepackage{amsbsy}
\usepackage{graphicx}
\usepackage{mathrsfs}

\newcommand{\tensor}[1]{\boldsymbol{\mathscr{#1}}}
\def\x{{\mathbf x}}

\definecolor{iccvblue}{rgb}{0.21,0.49,0.74}
\usepackage[pagebackref,breaklinks,colorlinks,allcolors=iccvblue]{hyperref}

\def\paperID{*****} 
\def\confName{ICCV}
\def\confYear{2025}

\title{Efficiently Generating Multidimensional Calorimeter Data with Tensor Decomposition Parameterization}

\author{Paimon Goulart$^*$\\
{\tt\small pgoul002@ucr.edu}
\and
Shaan Pakala$^*$\\
{\tt\small spaka002@ucr.edu}\\\\
University of California, Riverside
\and
Evangelos Papalexakis\\
{\tt\small epapalex@cs.ucr.edu}
}
\begin{document}
\maketitle
\renewcommand{\thefootnote}{\fnsymbol{footnote}}
\footnotetext[1]{Equal contribution}
\begin{abstract}
Producing large complex simulation datasets can often be a time and resource consuming task. Especially when these experiments are very expensive, it is becoming more reasonable to generate synthetic data for downstream tasks. Recently, these methods may include using generative machine learning models such as Generative Adversarial Networks or diffusion models. As these generative models improve efficiency in producing useful data, we introduce an internal tensor decomposition to these generative models to even further reduce costs. More specifically, for multidimensional data, or tensors, we generate the smaller tensor factors instead of the full tensor, in order to significantly reduce the model’s output and overall parameters. This reduces the costs of generating complex simulation data, and our experiments show the generated data remains useful. As a result, tensor decomposition has the potential to improve efficiency in generative models, especially when generating multidimensional data, or tensors.
\end{abstract}    
\section{Introduction}

Producing complex simulation data is a vital step to understanding physical systems, in domains such as climate science \cite{eyring2016overview,lam2023learning,ohana2024well} and high energy physics \cite{khattak2022fast, hep2019roadmap} . Unfortunately, performing the experiments to generate this simulation data sometimes requires excessive amounts of time and computational resources \cite{hep2019roadmap,ohana2024well}. For this reason, we see generative models, such as Generative Adversarial Networks (GANs) \cite{goodfellow2020generative} and Denoising diffusion models \cite{NEURIPS2020_4c5bcfec, song2022denoisingdiffusionimplicitmodels}, used for the generation of complicated simulation data \cite{khattak2022fast,amram2023denoisingdiffusionmodelsgeometry}. This is especially useful as these datasets have higher resolution, becoming more expensive to generate experimentally and using overwhelming amounts of computing power \cite{hep2019roadmap, khattak2022fast,amram2023denoisingdiffusionmodelsgeometry}.

To further accelerate this dataset generation, we introduce an internal tensor decomposition for these generative models, to significantly reduce the model parameters and increase overall efficiency. 
When generating multidimensional data, or tensors, we instead generate the much smaller tensor factors. Then after we generate these factors, we can combine them into the full generated tensor \cite{shiao2024tengan}. By doing this, we significantly reduce the number of parameters that are directly output by the model, reducing the overall model parameters as well. In our experiments, we show the viability of this internal tensor decomposition by comparing it with standard generative models that output the entire generated tensor. We make comparisons using the widely used Fréchet Inception Distance (FID) \cite{heusel2017gans}, to quantify how well the distribution of generated data matches the distribution of original data. In our results, we see that we can reduce the number of parameters by decreasing the tensor decomposition rank, and still maintain similar performance to the generative models with more parameters. Our code is available at \href{https://github.com/Pie115/GenTen-GAN-Diffusion}{https://github.com/Pie115/GenTen-GAN-Diffusion}.
\section{Preliminaries}

\subsection{Tensors \& Tensor Decomposition}

Tensors are multidimensional arrays. In other words, a vector is a 1-dimensional tensor, and a matrix is a 2-dimensional tensor. We will be looking at tensors of 3 or more dimensions \cite{kolda2009tensor,sidiropoulos2016tensor}.

\subsubsection{Tensor Decomposition}

Tensor decomposition is the process of expressing a tensor using smaller factors. For example, a common method of tensor decomposition is Canonical Polyadic Decomposition (CPD) 
\cite{kolda2009tensor,sidiropoulos2016tensor}. CPD expresses a tensor as a sum of rank-one tensors. A rank $R$ decomposition of a third-order tensor $\tensor{X} \in \mathbb{R}^{I \x J \x K} $ would be expressed as:
$
\tensor{X} \approx \sum_{r=1}^{R} (\mathbf{a}_r \circ \mathbf{b}_r \circ \mathbf{c}_r),
$
where $\mathbf{a}_r \in \mathbb{R}^I, \mathbf{b}_r \in \mathbb{R}^J, \mathbf{c}_r \in \mathbb{R}^K$, \text{ and } $\circ$ denotes outer product.

\subsection{Generative Adversarial Networks}

Generative Adversarial Networks (GANs) \cite{goodfellow2020generative} are a type of ML model that generates synthetic data. It utilizes two internal ML models: a generator and a discriminator. The generator is trained to generate realistic images based on a set of real data. The discriminator is trained as a binary classification model to distinguish between the real dataset and the generator's output. By training these two models in this way, the generator is encouraged to generate realistic images that will fool the discriminator.

\subsection{Diffusion Models}
Denoising diffusion models \cite{NEURIPS2020_4c5bcfec, song2022denoisingdiffusionimplicitmodels} are a class of generative ML models, which like GANs, are used to generate new data. Unlike GANs however, diffusion models do not need to perform adversarial training. Instead, a neural network is trained to reverse a process which denoises data that is corrupted with varying amounts of Gaussian noise. New samples are then produced by iteratively denoising pure Gaussian noise into a new image.
  \section{Methods}

\subsection{Tensor Decomposition in GANs}

\begin{figure}[!ht]

    \centering
    \includegraphics[width = 0.5\textwidth]{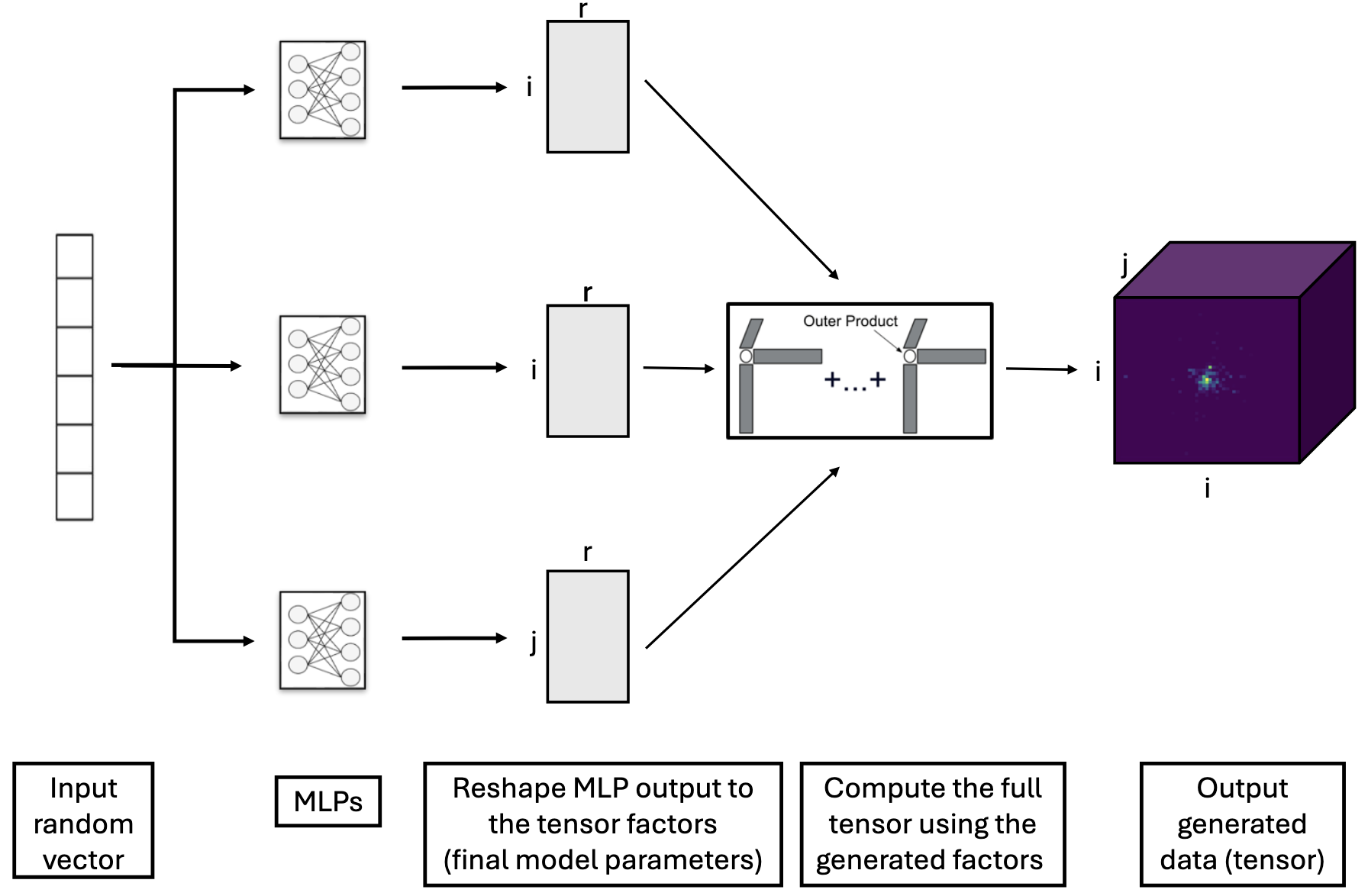}

    \caption{Example GAN with internal tensor decomposition. The GAN takes in a 1-dimensional random ($N(0,1)$ distribution) latent vector and passes it through MLPs to generate tensor factors. Then these tensor factors can be combined (using CPD in this figure) to produce the final generated tensor. In this figure, the generator produces three factor matrices of dimensions $\mathbb{R}^{i \times r}$, $\mathbb{R}^{i \times r}$, and $\mathbb{R}^{j \times r}$. Then we combine these factor matrices to produce the final generated tensor of shape $\mathbb{R}^{i \times i \times j}$. \label{tensor_GAN_figure}}

\end{figure}

Utilizing tensor decomposition in GANs allows it to generate multidimensional data, while requiring far less parameters. In Figure \ref{tensor_GAN_figure}, we visualize an example of a GAN using tensor decomposition to generate tensors.

The tensor in this example is of shape $\mathbb{R}^{i \times i \times j}$, so generating the full tensor would require $i^2 \times j$ output parameters. Generating the tensor factors \cite{shiao2024tengan} only requires $(i+i+j)\times r$ output parameters, which is usually much less.

\subsubsection{Generator}

The generator loss is the average discriminator's output on the generated data. Since the discriminator outputs a decimal value closer to 1 if it predicts generated data, and closer to 0 if it predicts real data, the generator wants to minimize this number. If the generator can minimize the discriminator's average predictions on the generated data (closer to 0), this means the discriminator is predicting real for the generated data. This hopefully means the generator is producing data that is indistinguishable from the real data.

\subsubsection{Discriminator}

For the discriminator, we take in the tensor as input and perform 2D convolutions on each slice. For example, the Calorimeter data is of shape $\mathbb{R}^{25 \times 51 \times 51}$, with 25 `slices' (each slice is a 2D matrix) and a height and width of 51. Then the discriminator pools these outputs and passes this through a dense layer and sigmoid layer to essentially perform logistic regression on the intermediate layer. If the output is closer to 1, then it is predicting the input is generated data, and if it is closer to 0 then it is predicting real data. For the discriminator training, we use PyTorch's Binary Cross Entropy loss function to evaluate the discriminator's ability to distinguish between the real and generated data.

\subsection{Tensor Decomposition in Diffusion Models}
In our model, we adopt a denoising diffusion implicit model (DDIM) framework. Compared to a standard denoising diffusion probabilistic model (DDPM), DDIM gives us faster sampling by using a deterministic non-Markovian process \cite{song2022denoisingdiffusionimplicitmodels}. Rather than requiring a full sequence of $T$ diffusion steps, DDIM allows us to generate high-quality samples using a reduced number of steps.

Since it has been shown that predicting the original sample $x_0$ is theoretically equivalent to predicting noise $\epsilon$ \cite{luo2022understandingdiffusionmodelsunified} we decided to use the former. Rather than having the model learn to predict noise $\epsilon$, we predict the clean sample $x_0$ from a noisy observation $x_t$. This choice seems intuitive in our setting, as the model ultimately produces factorized components of the original sample, and predicting $x_0$ allows for direct comparison of these components or their reconstruction.

In order to explore how diffusion could be applied with tensor decomposition, we decided to look into two different approaches. First is factor to factor where the model learns to directly denoise the factor matrices themselves. Second is tensor to factor where the model operates on full tensors and learns to output the factor matrices. Both of these approaches leverage CPD, but differ in how the denoising process works and where factorization occurs. In the following sections, we describe each method in detail.

\subsubsection{Factor to Factor}

\begin{figure}[!ht]

    \centering
    \includegraphics[width = 0.55\textwidth]{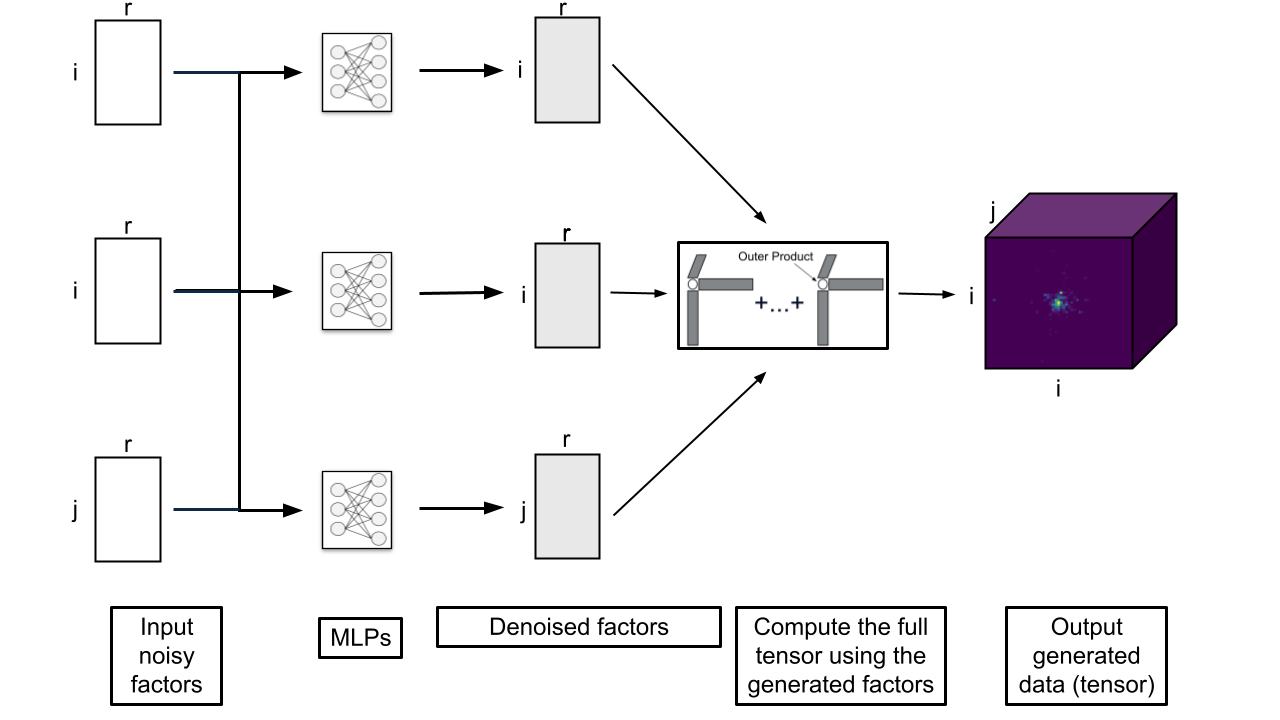}

    \caption{Factor to Factor diffusion sampling process. Each factor matrix is independently denoised through its own diffusion process, and the final tensor is obtained by combining the denoised factors (using CPD).\label{tensor_Diffusion_figure}}

\end{figure}

In order to fully leverage tensor decomposition with diffusion, instead of using one large model to generate an entire tensor $\tensor{X} \in \mathbb{R}^{I \times J \times K}$, one natural approach would be to split this process into three separate diffusion processes, each responsible for one factor matrix of a CPD. In Figure \ref{tensor_Diffusion_figure} we visualize this process, illustrating how each factor matrix is independently denoised in order to combine into one tensor sample. 

\begin{figure*}[!ht]

    \centering
    \includegraphics[width = 1.0\textwidth]{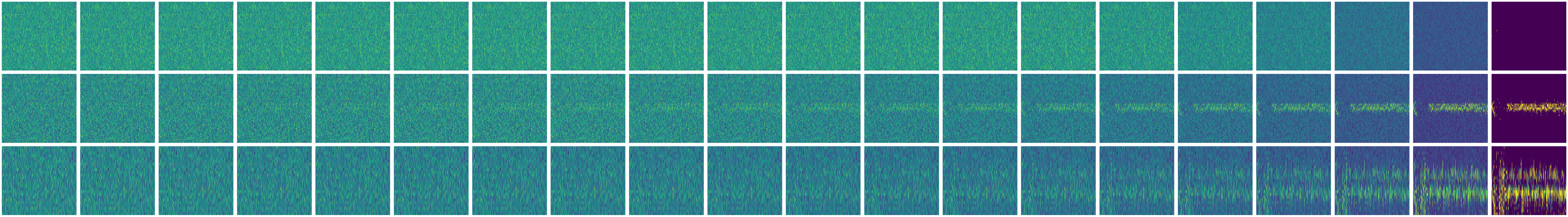}

    \caption{Progressive denoising of each factor matrix over time (from left to right). The three rows correspond to factor matrices A, B, and C, respectively (from top to bottom).\label{tensor_Denoising_Steps_figure}}

\end{figure*}

To achieve this, we first use CPD to decompose the tensors into three factor matrices. In this example, the factor matrices are of shape $a \in \mathbb{R}^{i \times r}, b \in \mathbb{R}^{i\times r}, c \in \mathbb{R}^{j \times r}$ where $r$ is the target rank. These three factor matrices are independently corrupted with varying amounts of Gaussian noise and are then used to train three models in parallel. During the sampling phase, we reverse this process. The trained models start from pure Gaussian noise and progressively denoise each factor matrix. In figure \ref{tensor_Denoising_Steps_figure} we visualize the denoising trajectory for each factor matrix.

In order to ensure that the models generate three factor matrices that correspond to one another, we calculate the loss as 

{\scriptsize
\begin{equation}
\mathcal{L}_{\text{total}} = 
\mathbb{E}_{t, A_0, B_0, C_0, \epsilon} \left[
\| A_{\text{pred}} - A_0 \|_F^2 +
\| B_{\text{pred}} - B_0 \|_F^2 +
\| C_{\text{pred}} - C_0 \|_F^2
\right]
\end{equation}}

In doing so, we significantly reduce both the input and output parameters when compared to operating on the full tensor space. From $i \times i \times j$ parameters to $(i+i+j)\times r$ parameters. Similarly to how latent diffusion models operate \cite{DBLP:journals/corr/abs-2112-10752} which perform diffusion in a compressed representation space. 

While this approach is intuitive and parameter efficient, one major drawback lies in the cost of decomposing the tensors ahead of time. At higher ranks, this decomposition becomes extremely computationally expensive.  As a result, despite reducing the number of model parameters, the overall pipeline can become less efficient. This motivates our second approach.

\subsubsection{Tensor to Factor}

\begin{figure}[!ht]

    \centering
    \includegraphics[width = 0.5\textwidth]{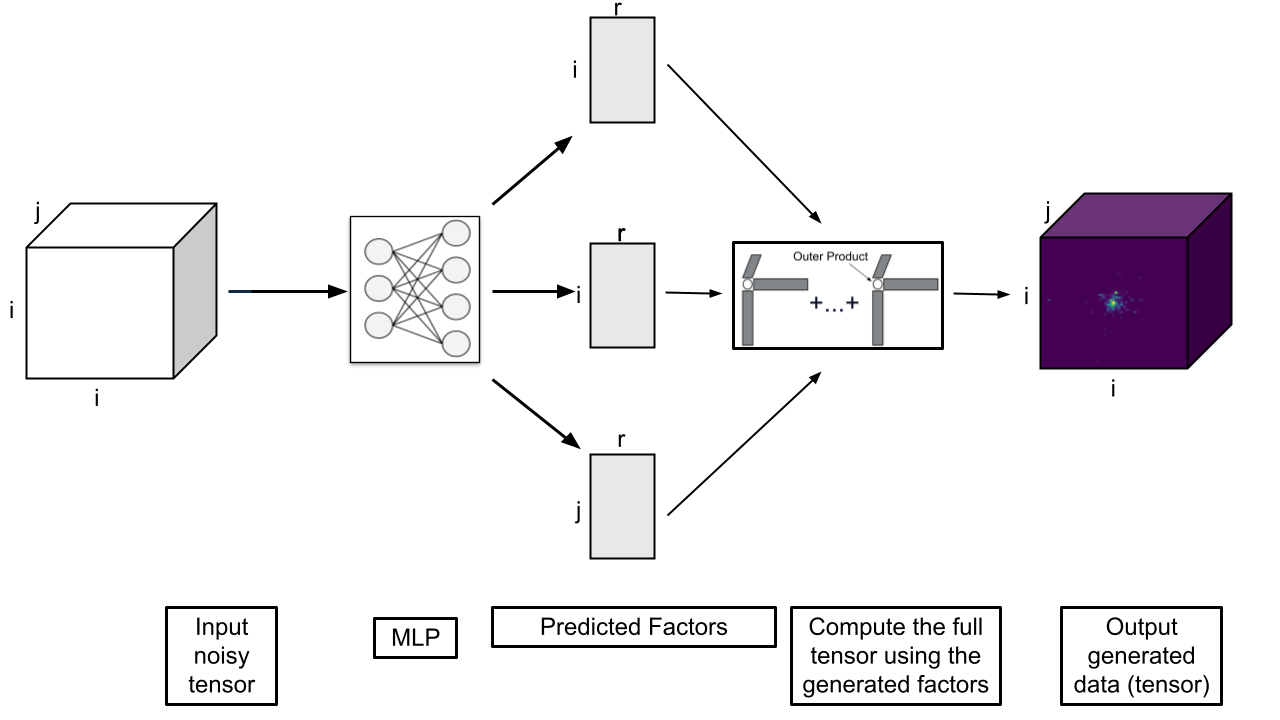}

    \caption{Tensor to Factor diffusion sampling process. A single model denoises a noisy tensor and predicts its corresponding factor matrices. These predicted factors are then combined using CPD to reconstruct the final tensor.\label{tensor_to_factor_figure}}

\end{figure}

To avoid the need to decompose tensors before training our model, we propose using a single model that operates directly on noisy tensors and learns to predict the three factor matrices that reconstruct them. In Figure \ref{tensor_to_factor_figure} we show this new process. This removes the overhead of decomposition while still reducing the output parameters.

To do this,  we begin with full tensors from the training dataset and corrupt them with varying levels of Gaussian noise. These noised tensors are then passed into a single model that outputs three factor matrices, denoted $a \in \mathbb{R}^{i \times r}, b \in \mathbb{R}^{i\times r}, c \in \mathbb{R}^{j \times r}$. In order to ensure that the model generates three factor matrices that accurately represent the original tensor, we first combine the three predicted factor matrices by taking the outer product $
\tensor{X}_\text{pred} \approx \sum_{r=1}^{R} (\mathbf{a}_r \circ \mathbf{b}_r \circ \mathbf{c}_r),
$ and compute the loss as

{\scriptsize
\begin{equation}
\mathcal{L}_{\text{total}} = 
\mathbb{E}_{t, \tensor{X}_0, \epsilon} \left[
\| \tensor{X}_\text{pred} - \tensor{X}_0\|_F^2
\right]
\end{equation}}

From this, we are able to maintain the benefit of reduced output parameters, but now without the need to decompose the tensors ahead of time. As a result, the model can be trained end-to-end directly on the data, making it more efficient and scalable in practice. 

\subsection{Evaluation}

For our evaluation, we choose to use the widely used Fréchet Inception Distance (FID) \cite{heusel2017gans, shmelkov2018good} to quantify how well the distribution of our generated data matches that of our real data.

\subsection{Dataset}

In our experiments, we use a publicly-available calorimeter simulation dataset as our multidimensional simulation data (\href{https://zenodo.org/communities/mpp-hep}{https://zenodo.org/communities/mpp-hep}). This data is essentially 3D images ($\mathbb{R}^{25x51x51}$) of electron \cite{ElectronShowerData}, neutral pion \cite{NeutralPionShowerData}, and photon \cite{PhotonShowerData} showers. 
\section{Experiments}

\subsection{Number of Parameters vs Performance}

\begin{figure}[!ht]
    \centering
    \includegraphics[width = 0.35\textwidth]{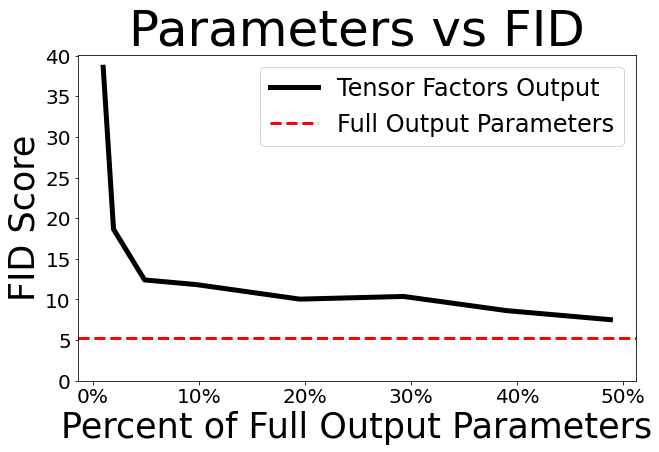}

    \caption{
    For the GANs, we display the Fréchet Inception Distance (FID) versus the output parameters. Output parameters shown as percent of full tensor (vs. smaller factor outputs). \label{gan_parameters_vs_fid}}

\end{figure}

\begin{figure}[!ht]

    \centering
    \includegraphics[width = 0.35\textwidth]{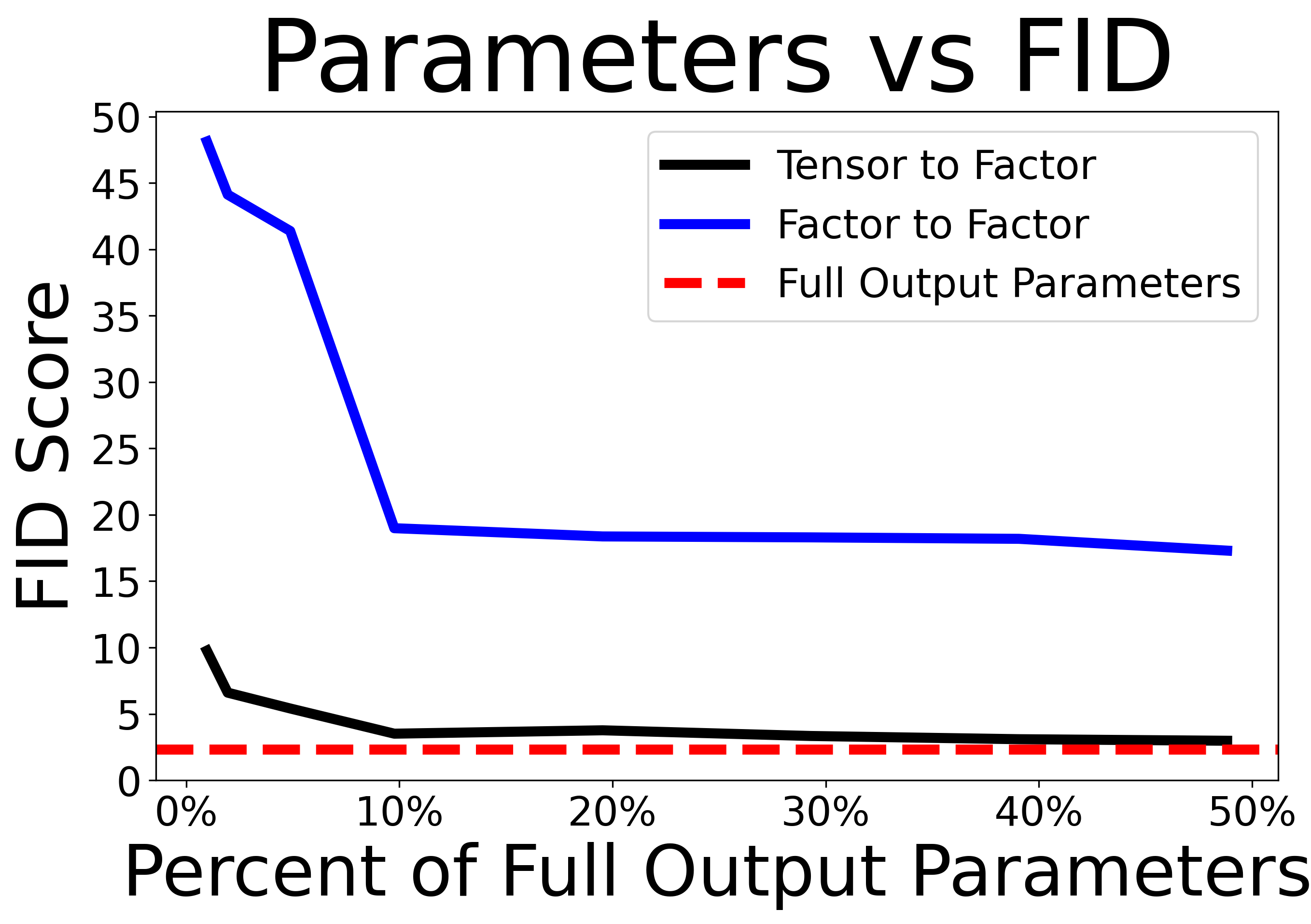}

    \caption{For the diffusion models, we display the Fréchet Inception Distance (FID) versus the output parameters. Output parameters shown as percent of full tensor (vs. smaller factor outputs).\label{diff_parameters_vs_fid}}

\end{figure}

In this experiment, we visualize the trade-off between performance and lower-rank tensor decompositions (less parameters). In Figures \ref{gan_parameters_vs_fid} \& \ref{diff_parameters_vs_fid}, we display the performance (using the Fréchet Inception Distance) of our models with different fractions of the full output parameters. We compare these with the performance of a GAN \& diffusion model with the full output parameters (red dashed line). For this comparison, we used a diffusion model without internal tensor decomposition. For the GAN comparison, we used a high-rank tensor decomposition such that the output parameters match the full tensor. Ideally, we would compare a GAN with no tensor decomposition, but this method outperformed other full-parameter GANs in our experiments.

For both GAN \& diffusion plots, we notice a sharp decrease in the FID, followed by a plateau at around 10-20\% of the full output parameters. This indicates that we can maintain similar performance (in terms of FID) while reducing parameters with a lower rank tensor decomposition.

For diffusion we observe that the tensor to factor variant achieves a remarkably lower FID than the factor to factor variant across all ranks. 
\section{Conclusion}

In these experiments, we show that tensor decomposition is promising for reducing the output size for generative models such as GANs and diffusion models. This is especially true as the output is multidimensional and higher resolution, as these simulation datasets often are. We are able to significantly decrease the output parameters and overall model parameters by utilizing tensor decomposition, and still generate useful synthetic data. 
\section{Acknowledgments}
Research was supported in part by the National Science Foundation under CAREER grant no. IIS 2046086 and grant no. IIS 1901379, by the Agriculture and Food Research Initiative Competitive Grant no. 2020-69012-31914 from the USDA National Institute of Food and Agriculture, and by the Army Research Office and was accomplished under Grant Number W911NF-24-1-0397. The views and conclusions contained in this document are those of the authors and should not be interpreted as representing the official policies, either expressed or implied, of the Army Research Office or the U.S. Government. The U.S. Government is authorized to reproduce and distribute reprints for Government purposes notwithstanding any copyright notation herein
{
    \small
    \bibliographystyle{ieeenat_fullname}
    \bibliography{main}
}

\newpage
\section*{Appendix}

\subsection*{Sample Images}

\begin{figure}[!ht]
    \centering
    \includegraphics[width = 0.5\textwidth]{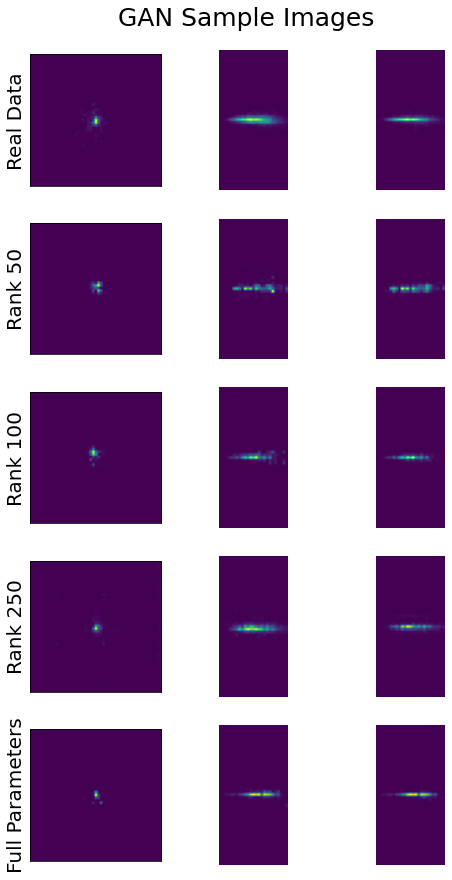}

    \caption{
    We display some sample images. The first row is the real data, the last row is a GAN using the full output parameters, and the remaining rows are GANs using various rank decompositions. For each row, we display different 2D "slices" of our 3D Calorimeter data. For our data of shape (i, j, k), we display 2D matrices of shape (i, j), (i, k), and (j, k). \label{gan_sample_images}}

\end{figure}

\begin{figure}[!ht]
    \centering
    \includegraphics[width = 0.5\textwidth]{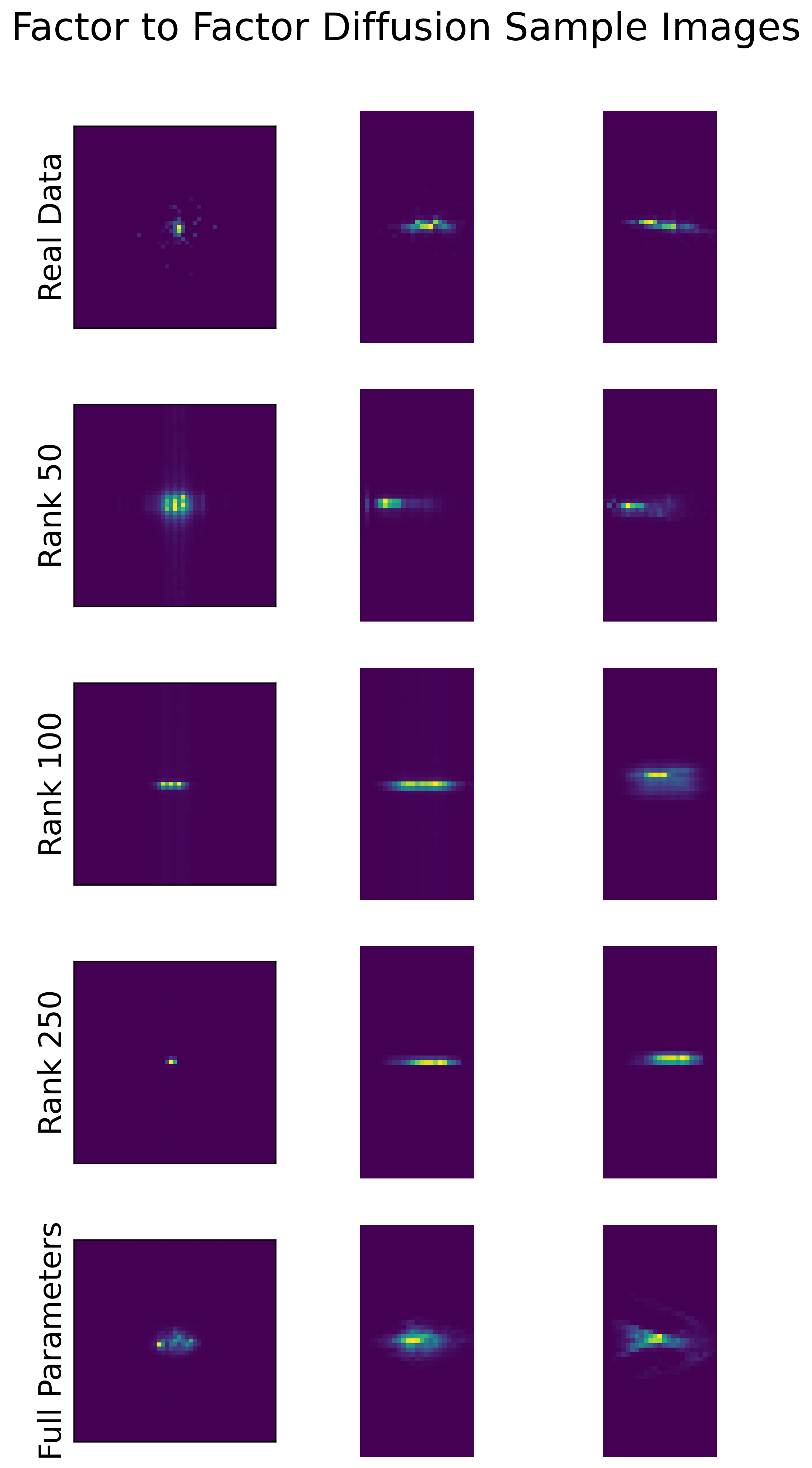}

    \caption{
    We display some sample images. The first row is the real data, the last row is a diffusion model using the full output parameters, and the remaining rows are factor to factor diffusion models using various rank decompositions. For each row, we display different 2D "slices" of our 3D Calorimeter data. For our data of shape (i, j, k), we display 2D matrices of shape (i, j), (i, k), and (j, k). \label{factor_to_factor_sample_images}}

\end{figure}

\begin{figure}[!ht]
    \centering
    \includegraphics[width = 0.5\textwidth]{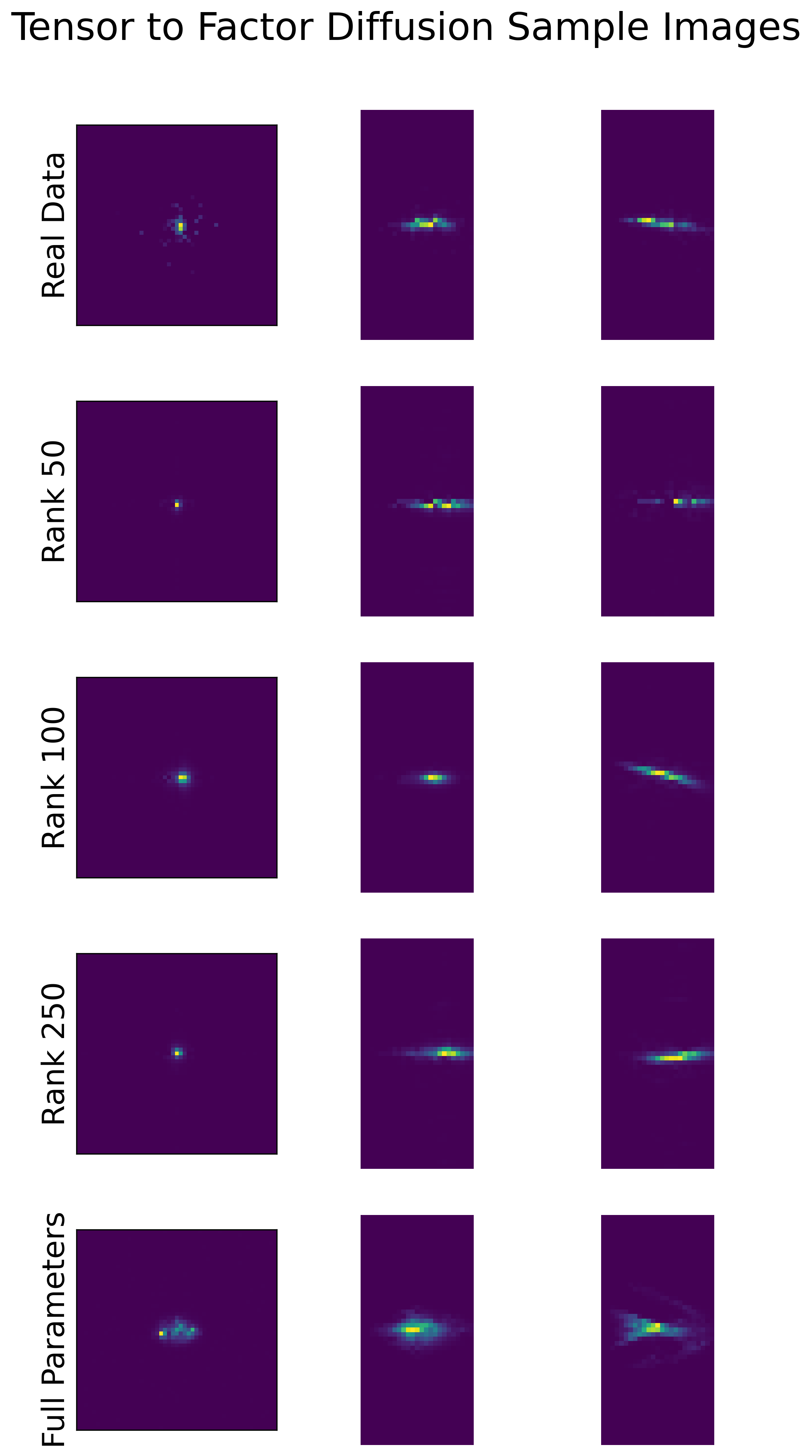}

    \caption{
    We display some sample images. The first row is the real data, the last row is a diffusion model using the full output parameters, and the remaining rows are tensor to factor diffusion models using various rank decompositions. For each row, we display different 2D "slices" of our 3D Calorimeter data. For our data of shape (i, j, k), we display 2D matrices of shape (i, j), (i, k), and (j, k). \label{tensor_to_factor_sample_images}}

\end{figure}

\subsection*{Future Work}

We plan to perform a wider variety of experiments and use different performance metrics to further demonstrate how useful the tensor decomposition is. This includes capturing the utility of the generated images by using it as additional training data in downstream ML tasks, as well as performing our experiments on a wider variety of datasets (e.g. higher order). In addition, we want to explore other tensor decomposition methods, such as Tucker \cite{balavzevic2019tucker}, to see if we can outperform CPD \cite{kolda2009tensor}.

\end{document}